\documentclass[11pt]{article}

\usepackage[preprint]{acl}

\usepackage{times}
\usepackage{latexsym}
\usepackage{booktabs}
\usepackage[T1]{fontenc}
\usepackage{todonotes}

\usepackage[utf8]{inputenc}

\usepackage{microtype}

\usepackage{inconsolata}

\usepackage{graphicx}

\title{Preference Optimization Improves Definition Generation: \\Semantic Change in EU Institutional Sustainable Finance Disclosures}

\author{Mariia Fedorova \\
  University of Oslo \\
  \texttt{mariiaf@ifi.uio.no} \\\And
  Andrey Kutuzov \\
  University of Oslo \\
  \texttt{andreku@ifi.uio.no} \\\And
  Gustavo Cadenas Delascio \\
  University of Oslo \\
  \texttt{g.c.delascio@globe.uio.no} \\}

\begin{document}
\maketitle
\begin{abstract}

Dictionary-like definition generators are known to be useful for lexical semantic change detection (LSCD). However, they suffer from overly granular sense distribution, producing excessive number of definitions for one and the same lexical sense. In this work, we show that fine-tuning with Direct Preference Optimization (DPO) helps to overcome this problem. We test our approach both quantitatively on a curated academic LSCD benchmark and qualitatively on real-world data for a practical task of auditing diachronic changes in sustainability terminology within project documentation of EU development finance. All models, code and predictions produced in this study are publicly available at \url{https://github.com/ltgoslo/dpo-for-definition-generation}.

\end{abstract}

\section{Introduction}

Diachronic lexical semantic change is a change of the meanings of words and phrases throughout time.  
While Average Pairwise Distance (APD) on target word embeddings \cite{giulianelli-etal-2020-analysing} is currently the recommended method of  change detection \cite{schlechtweg2025lscdbenchmarktestbeddiachronic}, generated dictionary-like definitions allow for clustering word usage examples by senses and labelling these word senses with human-readable explanations \cite{kutuzov-etal-2024-enriching}. This paves the way for explainable and interpretable semantic change modelling, potentially useful in practice for researchers from fields other than NLP.

Definition generators are large language models which undergo supervised fine-tuning (SFT) on datasets of target words used in specific contexts and human-produced definitions of target words in these contexts \cite{giulianelli-etal-2023-interpretable,fedorova-etal-2024-definition,tatarinov}.\footnote{E.g., PROMPT: `He sat on the bank of the river and cried. What is the definition of bank?', RESPONSE: `the rising ground bordering a lake, river, or sea or forming the edge of a cut or hollow'}

However, a known problem of definition generators is producing slightly different definitions for the same sense of a target word used in different contexts: the definitions may differ by only one word or even by one punctuation sign, while their semantics is the same or very similar. 
\citet{fedorova-etal-2024-definition} give an example of the English  word `\textit{plane}` being defined either as `\textit{An aircraft, especially one designed for military use}’ or `\textit{An aircraft, especially a military aircraft}’. These `noise senses` make it difficult for domain experts to analyse the resulting predictions and to draw conclusions about the nature of specific change cases. 

We approach this problem using preference optimization \cite{ouyang2022training}  of pre-trained definition generators. The general idea is that the desired behavior of the model is to generate very similar (or, ideally, identical) definitions for a target word used in the same sense. Thus, in our preference learning data, we choose one definition per sense as the preferred response and label other definitions (generated by the same model) as the rejected responses. Then we employ DPO \cite{rafailov2023direct} to push the model towards generating more uniform definitions per sense.

We find that fine-tuning definition generators with DPO does consistently reduce the number of definitions per target word (thus, the number of inferred senses), compared to the base models (which only underwent SFT). The picture is more nuanced for the performance on the graded semantic change benchmark used in SemEval'20 Task 1 \cite{schlechtweg-etal-2020-semeval}. Here, DPO can either improve or degrade the performance, depending on the base model family (see below).

Additionally, we conduct qualitative analysis of the performance of the resulting models on a domain-specific dataset of EU documents. We used a pipeline combining definition generation and the state-of-the-art XL-Durel model \cite{yadav-schlechtweg-2025-xl} to provide a domain expert a list of terms along with the dynamics of their senses changing over time.

Our contributions can be summarized as follows:

\begin{itemize}
    \item We show that preference optimization for less diverse definitions per sense improves the sense granularity of definition generators.
    \item It also improves the performance of the definition generator on semantic change detection, unless the model is already strong.
    \item We train a new family of definition generators based on OLMo-3-7B model, outperforming prior work based on Llama3-8B.
    \item We qualitatively evaluate the resulting models on the case of sustainability analysis on the corpus of European Investment Bank (EIB) documents, with a domain expert confirming our findings.
\end{itemize}

\section{Related work}

\paragraph{Semantic change} The task of diachronic semantic change detection (SCD) boils down to (given a set of $n$ historical corpora) either binary classification of a set of target words to either `changed' or `not changed', or to ranking of the target words by the degree of their semantic change (graded semantic change detection). One of the first standard sets for evaluating SCD was established in SemEval 2020 Task 1 \cite{schlechtweg-etal-2020-semeval}; we use its English subset to sanity-check that our definition generators do not perform worse on it than the ones from the previous work. Recently, contextualized definition generation was shown to be promising for modelling word meaning, and, subsequently, for tracing semantic change \cite{periti-etal-2025-definition}. This is the direction we pursue in this paper. However, the state-of-the-art approaches to semantic change use embedding vectors as sense representations. For example, XL-Durel is a word-in-context word sense disambiguation model, accepting a sentence with a specially tagged target word, e.g. \texttt{She sat on the <t>bank </t> of the river to rest}\footnote{\url{https://huggingface.co/sachinn1/xl-durel}}. The model's output are token embeddings of target words which are used as the APD input.

\paragraph{Definition generation} \citet{giulianelli-etal-2023-interpretable} and \citet{cassotti-etal-2024-using} pioneered employing definition generation for semantic change detection, with T5 encoder-decoder models. \citet{periti-etal-2024-automatically} fine-tuned decoder-only Llama 3 model on the same English data, excluding the test split\footnote{\url{https://hf.co/FrancescoPeriti/Llama3Dictionary-merge}}.
\citet{periti-etal-2025-definition} continued this line of work with English\footnote{\url{https://hf.co/FrancescoPeriti/LlamaDictionary-en_ML38BI}} and multilingual models trained on the DBnary dataset, consisting of Wiktionary\footnote{\url{https://en.wiktionary.org/wiki/Wiktionary:Main_Page}} data \cite{serasset-2012-dbnary}. \citet{fedorova-etal-2025-explaining} compared encoder-decoder and decoder-only models.

\paragraph{Sustainability analysis}
 Recent work establishes language models as a standard method for auditing corporate sustainability disclosures. For example, \citet{Luccioni2020} utilise RoBERTa-based question-answering models to extract climate-risk passages from heterogeneous financial reports, guided by specific disclosure frameworks. Similarly, \citet{Schimanski2024} develop specialised language models to classify corporate communication into distinct environmental, social, and governance categories, demonstrating a correlation between disclosure volume and external ESG ratings.  

\section{Datasets for supervised fine-tuning and preference optimisation}
\label{sec:datasets}

We use the same training and development splits of DBnary\footnote{\url{https://hf.co/datasets/FrancescoPeriti/Dbnary-en}. DBnary is licensed under Creative Commons Attribution-ShareAlike 3.0.} as \cite{periti-etal-2025-definition} and further divide them into SFT and DPO subsets (the SFT subset was only used for the OLMo-3 model). The splits were stratified by the number of unique gold senses. Their statistics are to be found in Table~\ref{tab:stat}. It is also worth noting that DBnary includes definitions both of single words and multi-word entities (`Leave behind', `Stroke of work' etc.).

An example of a SFT training instance is to be find in Appendix \ref{sec:app-sft}.

\begin{table}
\resizebox{\linewidth}{!}{%
\begin{tabular}{l|rr}
    \toprule
    \textbf{Split} & \textbf{Target words}& \textbf{Examples} \\
    \midrule
    SFT train & 90 030 & 220 325 \\
    SFT dev & 5 144 & 14 793 \\
    DPO train & 90 030 & 220 540 \\
    DPO dev & 5 145 & 14 407 \\
    \bottomrule
\end{tabular}
}
\caption{DBnary data split statistics before post-processing the DPO split.}
\label{tab:stat}
\end{table}

The DPO subset is used by the SFT-tuned models to produce chosen-rejected sample pairs as follows: we generate model's responses for all usage examples and post-process the resulting dataset so that `preferable' responses are always the same for usage examples of the same sense. `Preferable' definitions are chosen by the shortest Levenshtein distance
\cite{levenshtein1966binary} to the gold one. E.g., for the target word `\textit{juvenocracy}', the gold sense definition is `\textit{Government by youth}'. For three different usages of `\textit{juvenocracy}' in this sense, a definition generator produced the following three definitions: 1) `\textit{A government by youth or children}', 2) `\textit{Government by youth; government of children}', 3) `\textit{Government by youth; government by children or young people}'. Since definition 1 is most similar to the gold sense definition by edit distance, it is labelled as preferred; the rest of the definitions (2 and 3) are labelled as rejected, thus forming two preference learning pairs (1 vs 2 and 1 vs 3). After this post-processing, only senses with at least two usage examples are kept, with the resulting size of 20 016 examples in the train split and 2 086 in the development split. 

An example of a DPO training instance is to be find in Appendix \ref{sec:app-dpo}.

\section{Domain-specific corpus: EIB Environmental and Social Data Sheets}
In addition to evaluating our model on standard academic LSCD benchmarks, we also conducted qualitative analysis of their capabilities to trace sense changes in a domain-specific corpus. This section describes the corpus, for the analysis itself see Section~\ref{sec:eu_analysis}.

The EIB is the EU's biggest multilateral development bank and the main implementer of the Global Gateway strategy. Its lending decisions directly operationalise EU external investment policy at the project level. The Environmental and Social Data Sheet (ESDS) is a mandatory technical document produced for every EIB-financed project during the appraisal phase. It constitutes the environmental and social due diligence record submitted to the Board of Directors for financing approval  \citep{EIB2015}. 

Unlike press releases or public declarations, ESDS are operational instruments that influence capital allocation. Therefore, semantic changes within this corpus reflect actual bureaucratic implementation rather than political signalling. The corpus spans from 2014 to 2026. The starting point is EU directive 2014/52/EU and the Aarhus Regulation EC N°1367/2006, which regulate the assessment of projects on the environment and prompted the EIB to establish a public register. This window establishes the historical baseline of EIB sustainability discourse before capturing the 2021 launch of the Global Gateway and the concurrent repositioning of the EIB as a 'climate bank'. This allows us to distinguish semantic shifts associated with specific geopolitical mandates from broader baseline drift in sustainability vocabulary.

The ESDS collection is scraped from EIB website\footnote{\url{https://www.eib.org/en/registers/all/index}. The website's terms of use \url{https://www.eib.org/en/terms-of-use.htm} allow downloading data, but  are not clear about republishing/creating derivatives for research purposes. Thus, we will only publish the code to reproduce our processing, but not the collection itself.} (4 201 documents). Since language tags on the website has turned out to be not reliable, we apply language identification on the document level using OpenLID-v3 model\footnote{\url{https://hf.co/HPLT/OpenLID-v3}} \cite{burchell-etal-2023-open,fedorova-etal-2026-openlid}. 4 189 documents are in English, with the rest of 12 documents being in Albanian, Greek, and French. Thus, our study has to  be limited to English only. The documents are split into sentences by punctuation marks, resulting in 151 041 sentences. These are used as word usage examples for LSCD. 

We apply \texttt{flair/chunk-english}\footnote{\url{https://huggingface.co/flair/chunk-english}} \citep{akbik2018coling} model to extract all noun phrases in the corpus. Out of those, a domain expert samples target phrases of interest. This sample is used for qualitative analysis in Section~\ref{sec:eu_analysis}.

\section{Experiments}

\subsection{Baselines}

For evaluation on SemEval'20 Task 1, we use the following baseline models, all of which underwent SFT on definition generation for English: 

\begin{itemize}
    \item mT0-xl (4 billion parameters) \cite{fedorova-etal-2024-definition}
    \item Llama3Dictionary (8 billion parameters) \cite{periti-etal-2024-automatically}
    \item LlamaDictionary-en\_ML38BI (8 billion parameters) \cite{periti-etal-2025-definition}
\end{itemize}

The first two models were fine-tuned on a mixture of WordNet \cite{ishiwatari-etal-2019-learning}, Oxford \cite{gadetsky-etal-2018-conditional}, and CoDWoE \cite{mickus-etal-2022-semeval} datasets, while the third one was fine-tuned on DBnary. All of them were successfuly used in prior work on definition generation and semantic change modeling.

\subsection{Our definition generators}


We extend prior work by creating three new definition generators. First, we preference-optimize LlamaDictionary-en\_ML38BI model. This was an obvious experiment in order to test whether DPO can improve a definition generator. However, one downside of Llama-based LLMs is the lack of full openness: the training data of Llama models is unknown, and the model license is restrictive. Thus, we also used a more recent and fully open language model of about the same size: Olmo-3-7B-Instruct-SFT\footnote{\url{https://hf.co/allenai/Olmo-3-7B-Instruct-SFT}} \cite{olmo2025OLMo3}.

The original Olmo-3-7B-Instruct-SFT (7 billion parameters) was not specifically fine-tuned to generate contextualized definitions, so we first ran on it a round of SFT, using the corresponding DBnary split (see Section~\ref{sec:datasets}). After that, it was preference-optimized in the same way as LlamaDictionary-en\_ML38BI.

Thus, the three models we introduce here are Llama-DPO, OLMo-SFT and OLMo-DPO.

\subsection{Hyperparameters and hardware}
We used HuggingFace \texttt{trl}\footnote{\url{https://huggingface.co/docs/trl}} library for both SFT and DPO. All the DPO fine-tuning was done in one epoch using LoRA \cite{hu2022lora} and bfloat16 (bf16) precision to reduce the required compute. The SFT round on OLMo-3 was run on maximum 20 epochs (in practice, early-stopped after epoch 2). Table \ref{tab:hype} presents the hyperparameters of both SFT and DPO.

\begin{table}
\resizebox{\linewidth}{!}{%
\begin{tabular}{l|cc}
    \toprule
    \textbf{Parameter} & \textbf{SFT} & \textbf{DPO} \\
    \midrule
    LoRA rank & 256 & 64 \\
    LoRA alpha & 512 & 16 \\
    LoRA dropout & 0.1 & 0.2 \\
    learning rate schedule & linear & linear \\
    learning rate & 2e-5 & 1e-6 \\
    batch size & 32 & 24 \\
    \bottomrule
\end{tabular}
}
\caption{Training hyperparameters.}
\label{tab:hype}
\end{table}

Note that our LoRA alpha and rank parameters for DPO are much lower than for SFT, since at this step the model  should change only slightly. We also found that too aggressive DPO (when using $\beta$ in the range 0.4-0.1) increases predictions' diversity rather than declines it (we track the average number of sense per word on the development data split during training), thus we use the value 0.5.

DPO was done on a single NVIDIA GH200 GPU for Llama3-based models, and on two GPUs for OLMo-3 based ones\footnote{It is absolutely possible to fine-tune OLMo-3 on a single GPU as well.}. 1 epoch of DPO took not more than 2 hours. SFT was done on two NVIDIA GH200 GPUs, with 5 hours walltime.

\subsection{Change detection}
\label{subsec:lscd}

The definition generators described in the previous sections were used to produce contextualized definitions for selected occurrences of target words in SemeEval'2020 English corpora (historical periods 1810–1860 and 1960–2010). The generated definitions serve as interpretable representations of the target words' semantics in context. We fully re-use the `definitions as senses' change detection algorithm proposed in \cite{fedorova-etal-2024-definition},  employing Jensen-Shannon divergence (JSD) to compare sense distributions across time periods, and applying definition generators to exactly the same usage examples. 

Our hypothesis is that preference-optimized definition generators will (at least somewhat) alleviate the issue of redundant definitions per sense. 
\citet{fedorova-etal-2024-definition} already tried to address this issue by merging together definitions with low Levenshtein edit distance between them. But they could not tune the merging threshold hyper-parameter because of the lack of development dataset. In this work, we use the DBnary development split for this purpose, so  we tune merging threshold on it (optimizing the number of senses per target word), before final evaluation on the SemEval'2020 test set. The resulting optimal threshold with full merging was found to be 22, and we refer to it below as `tuned threshold'.

Since we expect that a surface metric like an edit distance would not capture cases where two definitions express the same meaning using different words, we also tried Agglomerative Clustering \cite{kaufman2009finding} of definitions' embeddings produced by \texttt{BAAI/bge-m3} model \cite{bge-m3}\footnote{\texttt{BAAI/bge-m3} is a sentence encoder for information retrieval. It accepts sentences `as is', without any preprocessing, as an input. } with the complete linkage and cosine distance threshold 0.2. 

Finally, for the qualitative analysis in Section~\ref{sec:eu_analysis}, we re-rank the target words according to their change score using APD on the embeddings obtained\footnote{We used the lemmatizer based on HPLT 3.0 \cite{oepen2025hplt30largescalemultilingual} BERT model (\url{https://hf.co/HPLT/hplt_gpt_bert_base_3_0_eng_Latn-UD}) for finding the target words in the usage examples.} from the state-of-the-art word-in-context model XL-Durel \cite{yadav-schlechtweg-2025-xl}, so that the domain expert gets true positive changes on top of their list.\footnote{This model was trained to produce sentence embeddings of usage examples with a target word surrounded by special symbols. We were able to reproduce XLD-APD performance from Figure 5 in \cite{schlechtweg2025lscdbenchmarktestbeddiachronic}, achieving the score of 0.768.} The mixture of APD and `definitions-as-senses' lets the user not only see which words have changed, but also which exact senses were gained and which ones were lost. 

Predictions by \texttt{BAAI/bge-m3} and XL-Durel run on a single Google Colab's T4 GPU in not more than 15 minutes.

\subsection{Results}

\begin{table*}
\resizebox{\linewidth}{!}{%
\begin{tabular}{l|cccccc}
    \toprule
    \textbf{Model} & \textbf{mT0-xl SFT} & \textbf{Llama3Dictionary SFT}  & \textbf{LlamaDictionary-ML38B SFT} & \textbf{OLMo-3 SFT} & \textbf{LlamaDictionary-ML38B DPO} & \textbf{OLMo-3 DPO} \\
    \midrule
    No merging & 0.476 [0.18, 0.69] & 0.429 [0.12, 0.66] & 0.517 [0.23, 0.72] & \textbf{0.586} [0.32, 0.76] & 0.532 [0.25, 0.73] & 0.546 [0.27, 0.74] \\
    \midrule
    Minimalist merging 50  & 0.565  & 0.547  & 0.237  & 0.445  &  0.354 &  0.258\\
    Full merging 50  & 0.493   & 0.494   & 0.200 & 0.389 & 0.224 & 0.170   \\
    Full merging 22 (tuned) & 0.431 & 0.510 & 0.444 & 0.562  &0.426 & 0.380 \\
    \midrule
    Clustering & 0.435 & 0.465 & 0.464 & 0.557 & 0.493 & 0.498 \\
    Clustering separately & 0.171 & 0.214 & 0.42 & 0.207 & 0.458 & 0.068 \\
    \bottomrule 
\end{tabular}
}
\caption{Evaluation on SemEval'20 Task 1, Spearman rank correlations. Generated definitions were obtained in single runs, and sense detection algorithms are deterministic. Confidence intervals for the best performing postprocessing (no postprocessing) are reported in brackets. 
}
\label{tab:results}
\end{table*}

Table~\ref{tab:results} shows Semeval 2020 Task 1 subtask 2 (graded semantic change detection) results for baselines and our models. We only focus on subtask 2, because it better suits our research problem of finding change of non-discrete senses in the ESDS data. In this work,  we report the results with no merging, with the merging parameter from \cite{fedorova-etal-2024-definition} (threshold 50), with the optimal merging threshold tuned on DBnary development set (22, full merging), and with two strategies of clustering (done separately for each time period -- we tried it because in the original paper it was shown to perform better -- and on all available predictions for both time periods).

One should note that Semeval 2020 Task 1 is a small dataset: it only contains 37 target words. As \citet{Phan-Tat-2026} point out, this limitation makes it impossible to fairly rank the performance of different systems outputting close scores, because these scores yield overlapping confidence intervals (CIs). Following this work, we have calculated CIs for the scores of our models. We find no consistent improvement from merging or clustering across models, training data and fine-tuning approaches; the best absolute score is achieved without any postprocessing. The models fine-tuned on DBnary perform slightly better than those fine-tuned on the Wordnet-Oxford-CoDWoE mixture following \cite{giulianelli-etal-2023-interpretable}. However, these scores do not allow to conclude whether OLMo or LlamaDictionary-ML38B is a better base model choice, or whether the DPO step is necessary, since their scores are within the intersection of their CIs. Overall, definition generation does not outperform sentence embedding approaches such as XL-Durel+APD (either because of high granularity or because word usages representing some senses were left behind in our samples); but our models have passed the sanity check making them preferrable over previous work.

Table~\ref{tab:senses} gives more details on the influence of DPO on  definition generation models. We use DBnary development set (5 145 target words, 14 407 usage examples) as a human performance baseline to compare our models against. Specifically, we count the average number of senses (unique definitions) per target word, the ratio of ambiguous target words and the average number of senses per \textit{ambiguous} target word. 

Distribution of the senses in DBnary and models' predictions differs a lot. First, we see that definition generators overestimate the number of senses per word and the ratio of ambiguous words (i.e., they yield more than one sense for some mono-semantic words). Applying DPO reduces number of predicted senses, confirming our original hypothesis. The average number of senses per ambiguous word is mirroring this dynamics (if we are keeping the total number of senses fixed and increasing the share of ambiguous words, the average ambiguity of these words goes down). Thus, DPO does push the models towards the gold DBnary distribution.

The ability to reproduce sense distribution in one dataset may not directly translate to higher performance in semantic change detection on another dataset. OLMo-3 models are much more prone to generating diverse definitions than Llama models: the percentage of ambiguous words is more than twice higher than in gold DBnary distribution.

\begin{table}
\resizebox{\linewidth}{!}{%
\begin{tabular}{l|lccc}
    \toprule
\textbf{Model} & \textbf{Total senses} &\textbf{Senses per word} &\textbf{\% amb.} & \textbf{Senses per amb.}  \\
\midrule
Human (DBnary gold) & 7 645  & 1.49 &  20.1 & 6.76  \\
\midrule
Llama ML38B SFT & 9 228 & 1.79 & 30.2 &  5.50  \\
Llama ML38B DPO  & 9 044 & 1.76 & 29.3 & 5.58  \\
OLMo-3 SFT  & 12 721 & 2.47  &46.7 & 4.68  \\
OLMo-3 DPO  & 12 143 & 2.36 & 45.3 & 4.72  \\
\bottomrule
\end{tabular}
}
\caption{Distribution of senses in the DBnary development set and in the predictions of our models. `Amb.' stands for `ambiguous words' (more than one sense).
}
\label{tab:senses}
\end{table}

\section{Case study: qualitative sustainability analysis}
\label{sec:eu_analysis}
In sustainability studies, core terminology is highly contested and semantically ambiguous \citep{Connelly2007}. Policy institutions selectively leverage these elastic definitions to classify investments and justify environmental action.  By stretching definitions to satisfy conflicting pressures, institutions maintain strategic ambiguity \citep{Leitch2007}. This ambiguity generates a significant divergence between an institution's discursive commitments and its operational behaviour \citep{Fiss2006}. Consequently, ambitious sustainability mandates often produce implementation gaps where narrative outpaces actual financial additionality.

The EU's Global Gateway strategy illustrates this dynamic. Launched in 2021 as a geopolitical infrastructure investment strategy, it repositioned existing EU development cooperation pipelines under a 'green transition' narrative. However, quantitative analyses reveal limited new capital allocation, suggesting a relabelling of existing commitments \citep{Bilal2024}. Detecting whether this shift reflects a genuine semantic evolution in operational definitions or merely a strategic redeployment of terms requires computational methods capable of resolving fine-grained sense distinctions over time.

To assess whether the semantic evolution in EU project documents indicates genuine change in meaning or strategic ambiguity, we apply our definition generation pipeline (with both SFT-only and DPO models) to the EIB corpus. We qualitatively evaluate both the scalar detection of change and the granularity of the generated definitions from a domain-expertise perspective, comparing the models across two primary axes: the base model (Llama vs. OLMo-3) and the training method (SFT vs. DPO). The models' output was first pre-filtered using APD, as described in Subsection~\ref{subsec:lscd}.

\subsection{Validating Change Detection via APD}
We first evaluate the alignment of the APD scores with expected policy shifts. The APD ranking consistently separates stable institutional vocabulary from highly contested policy terms. Terms expected to remain stable, such as `public procurement', `private company', and `environmental NGO', are correctly assigned low APD scores (e.g., $<0.09$).
Conversely, the terms subjected to recent regulatory shocks are assigned high APD scores. For example, `environmental documents' exhibits significant semantic distance when comparing 2016 and 2021 occurrences. This computationally detected shift aligns with the 2020 publication of the EU Taxonomy, which fundamentally altered the regulatory requirements for environmental documentation. Similarly, emerging technologies like `new hydrogen' return high APD scores ($0.391$), accurately reflecting ongoing debates regarding sustainable fuel classifications. Based on domain validation, we observe an empirical APD threshold of approximately $0.27$ separating substantive policy shifts from baseline syntactic variation. Only terms with the APD change score exceeding this threshold were qualitatively analyzed in the subsections below.

\subsection{Sense Granularity and Preference Learning}
While scalar metrics detect change, definition generation explains its nature. To evaluate this, a domain expert analysed a subset of 170 candidate phrases with sufficient temporal coverage (1,666 candidate pairs), coding each generated definition against a baseline sense as unchanged, hallucinated, or into one of three inductively derived categories of semantic shift (detailed in Section 6.3): financialisation, normative stretching, or regulatory absorption. 53\% of pairs show no meaningful change, and 15\% could not be reliably compared due to mismatched phrase boundaries. Among the remaining pairs, 21\% are hallucinations, 20\% reflect the reappearance of a previously attested sense (flagged as reintroductions rather than novel gains), 19\% financialisation, 16\% normative stretching, 12\% regulatory absorption, and 12\% combine multiple types. Qualitative inspection reveals stark differences in how models manage this sense granularity. 
Evaluating the training method axis, the LlamaDictionary-en\_ML38BI SFT baseline suffers from excessive sense granularity. It frequently generates entirely new definitions (senses) for minor syntactic variations rather than genuine semantic shifts, producing 130 distinct senses for 'quality management' and differentiating 'protecting the environment' from 'synonym of environmental organisation' for 'environmental NGO'. Furthermore, it generates hallucinations for standard technical terms, such as linking 'necessary activity' to the US Constitution. Applying DPO to the LlamaDictionary-en\_ML38BI model reduces this granularity and, when paired with clustering, consolidates terms into more stable senses while mitigating some hallucinations.
Evaluating the architecture axis, the OLMo-3 models produce much lower sense granularity than Llama. OLMo-3 SFT performs exceptionally well in structural coherence and exhibits the lowest hallucination rate of all tested models. OLMo-3 DPO achieves the lowest granularity, successfully reducing redundancy. However, as detailed below, this structural advantage comes at a severe cost to domain utility.
\subsection{Tracking Semantic Shifts in Policy Implementation}
Crucially, the LlamaDictionary-en\_ML38BI DPO model exhibited the highest sensitivity to policy-relevant nuance, successfully mapping strategic semantic expansion in the EIB corpus where other models failed. As no established typology fits this institutional-finance context, the domain expert inductively derived three trajectories of semantic shift from the generated definitions, informed by the literature on strategic ambiguity: the financialisation of physical metrics, the normative stretching of terms to absorb ecological premiums, and the regulatory absorption of external mandates into existing vocabulary. These shifts manifest as follows:

\begin{itemize}
\item \textbf{Financialisation:} 'additional CO2 emissions' shifted from a strict physical sense in 2014 ('additional carbon dioxide') to a financial one by 2016 ('additional cost of carbon dioxide'); similarly, 'average lifetime' moved from describing physical objects in 2019 to financial entities by 2022 ('the life of an asset'). 
\item \textbf{Normative stretching:} 'broad participation' expanded from a purely quantitative baseline in 2017 ('participation by a large number of people') to an explicitly political one by 2022 ('involvement of as wide a section of a community as possible'). Furthermore, 'available techniques' showed a generic manufacturing practices sense in 2024 but one about methods specifically mitigating environmental damage in 2017.
\item \textbf{Regulatory absorption:} 'ambitious target' shifted from a generic sense in 2015 ('a challenging goal that is striven for') to an EU-specific one by 2020 ('a goal to cut a particular environmental pollutant by a specific date').

\end{itemize}

Phrases can also acquire more than one shift type at once: 'ambitious target' independently developed a normative reading by 2023 ('an ambitious goal to reduce pollution or increase in environmental quality') alongside its regulatory one, without losing its original generic sense, which illustrates the strategic ambiguity that institutions rely on to satisfy conflicting mandates.

\subsection{Observed Limitations in Definition Generation}
Qualitative analysis exposes a clear trade-off between structural granularity and domain sensitivity across both comparison axes.
First, comparing architectures, the OLMo-3 models sacrifice critical policy sensitivity to achieve their low granularity. While Llama models capture nuanced shifts in terms like 'additional CO2 emissions', OLMo-3 models generate overly generic, dictionary-style definitions that fail to capture institutional context.
Second, comparing training methods within architectures reveals diverging effects. For Llama, DPO is a necessary intervention to make the highly granular SFT outputs usable, preserving policy sensitivity while smoothing structural noise. Conversely, applying DPO to OLMo-3 damages its qualitative performance. While OLMo-3 SFT maintains some alignment with the EIB context and minimizes hallucinations, OLMo-3 DPO collapses specialised terms entirely. For instance, OLMo-3 DPO hallucinates 'new hydrogen' as a 'Japanese fuel' and produces definitions entirely detached from the political economy context. This indicates that while preference optimisation improves highly granular models like Llama, it excessively penalises the domain-specific jargon required for institutional analysis in models like OLMo-3.
\subsection{Bridging Detection and Interpretation}
Our qualitative assessment shows that preference-optimised definition generation successfully translates abstract semantic change scores into explicit, human-readable evidence. While scalar metrics flag that a change has occurred, our pipeline maps these contextualised occurrences to coherent definitions-as-senses, providing political scientists and sustainability researchers with a scalable tool to track how operational terms related to green investments evolve across institutional disclosures. Ultimately, by complementing change scores with these structured definitions, researchers and policymakers can systematically test claims such as the Global Gateway's 'green transition' narrative against project-level operational vocabulary, distinguishing genuine change in meaning from an institution's strategic expansion of core concepts, checkable at corpus scale rather than through manual document review.

\section{Conclusion}

We have shown that preference optimization (in the form of DPO) may help to overcome the problem of too granular senses in the outputs of existing definition generation models. This in turn increases the performance of semantic change detection systems in the `definitions as senses' paradigm. 

We experimented with different strategies of merging definitions-as-senses where they were still too granular and found that it is largely not necessary for modern definition generators. 

Interestingly, a large crowd-sourced dictionary (DBnary) allows to train more powerful models than datasets derived from smaller  dictionaries created by professionals. Another practically useful finding is that a 7 billion parameter model performs on par with a 8 billion parameter one, although it is not possible to name a reason for it from our study (the models are also different in their pretraining and instruction-tuning data, tokenizers etc.).

While the gain in reducing number of senses may look minor in absolute numbers compared to the computational cost of DPO, for a human end user it means much less (may be hundreds) definitions to read and analyse, so we still believe DPO is useful. Preference optimization has to be done only once and we have already done it. Any social scientist will be able to use the models we publish. 

We applied our pipeline to a real-life task of studying diachronic change 
of sustainability concepts in EU project documents. Applied to the EIB corpus, the pipeline made visible how the Bank's own operational vocabulary is reclassified over time to adapt to external political and regulatory mandates. It offers concrete evidence that a definition generator retaining domain-specific nuance, as achieved here by LlamaDictionary-en\_ML38BI DPO, can make an institution's narrative shifts traceable directly in its operational documents, not only its curated public statements.

\section*{Limitations}

A potential risk of our work is that we did not check DBnary for offensive content (while Wiktionary is moderated, it can be edited by anyone); we also did not check the base models for the ability to generate such a content.

Our experiments with clustering are limited to only one sentence embedding model and to only one algorithm. However, we believe that the future work should concentrate on developing better definition generators rather than on dataset-specific post-processing of the definitions.

We do not use any thresholding for APD. Thus, our current pipeline does not map the exact values of APD to the `real' change/not change labels. This decision is currently left to a human expert.

Impressions of a human expert do not exactly align with the models' performance on a curated benchmark (featuring only 37 words, all of them belonging to the basic vocabulary of the English language). The problem of LSCD models performing differently on curated benchmarks and `in the wild' has also been  reported in the prior work \cite{umarova-etal-2025-current}. Thus, there is a need for larger and domain-specific benchmarks. Qualitative categorisation in Section ~\ref{sec:eu_analysis} was conducted by a single domain expert; adding independent annotators and reporting inter-rater agreement is a natural next step for this line of work.

\section*{Acknowledgments}
We are thankful to Jonas Timmann Mjaaland for the help in software implementation.

The computations were performed on resources provided by Sigma2 - the National Infrastructure for High-Performance Computing and Data Storage in Norway.

This project has received funding from the European Union's Horizon Europe Marie Skłodowska-Curie Actions Doctoral Network under grant agreement No 101120060.

\bibliography{anthology,custom}

\appendix

\section{APD descriptive statistics}
\label{sec:appendix}

Table \ref{tab:stat-apd} presents descriptive statistics of APD calculated by XL-Durel for the ESDS data.

\begin{table}[h]
\centering
\begin{tabular}{l|c}
    \toprule
     & APD value \\
    \midrule
    mean & 0.13 \\
    std & 0.10 \\
    min & 0.0 \\
    25\% & 0.07 \\
    50\% & 0.11 \\
    75\% & 0.17 \\
    max & 0.72 \\
    \bottomrule
\end{tabular}
\caption{Descriptive statistics of APD calculated by XL-Durel.}
\label{tab:stat-apd}
\end{table}

\section{Examples of fine-tuning instances}

\subsection{SFT}
\label{sec:app-sft}

An example of a SFT training instance (used for fine-tuning the OLMo base model; we did not perform own SFT of \texttt{LlamaDictionary-en\_ML38BI}, but reused the prior work).
\begin{itemize}
    \item Target word: Suffer
    \item Prompt + usage example: If you keep partying like this, your school-work will suffer. What is the definition of suffer?
    \item Gold prediction: To become worse. 
\end{itemize}

\subsection{DPO}
\label{sec:app-dpo}

The next example shows a DPO training instance, created from \texttt{LlamaDictionary-en\_ML38BI SFT} predictions:

\begin{itemize}
    \item Target word: Tip 
    \item Prompt + usage example: Please provide a concise definition for the meaning of the word "Tip" in the following sentence: (Dirks was an investment analyst who learned from a former employee of Equity Funding that the company had been fraudulently manufacturing insurance policies. Dirks tipped several institutions which then liquidated \$16 million in Equity Funding stock before the fraud was exposed and the bottom fell out of the market. According to the Court, Dirks was a hero for (eventually) exposing the fraud. The SEC\'s censure of Dirks for tipping inside information was reversed because, the Supreme Court held, the liability of a tippee derives from that of the tipper. If the tipper is without sin, so is the tippee. Here, the employee\'s purpose in informing Dirks was to expose the fraud, so the tipper breached no duty. Moreover, according to the Court, for there to be a breach of duty, there must be personal benefit (in the form of pecuniary gain or reputational benefit that will translate into future earnings) arising to the tipper from the disclosure
    \item Chosen: To give a piece of private information to; to inform (someone) of a clue, secret knowledge, etc.
    \item Rejected: To give a piece of private information to; to inform (someone) of something not widely known.
\end{itemize}

In this instance, the chosen definition was generated from the shown usage example `(Dirks was...', and the rejected definition was generated from the usage example `If the institutional trading patterns we observe are a result of tipping, rather than precipitated by some other event, then we should see only a slight increase in the number of institutions active in the market (tipping would precipitate entry by, at most, the few institutions that were tipped). At the same time, given the nature of the reports, we should see an increase in the average buying activity of institutions (we cannot identify the specific institutions that were tipped, so we can only look at averages) and see little change in selling'. The original gold definition for both usages was `To give a piece of private information to; to inform (someone) of a clue, secret knowledge, etc. [from late 19th c.'.

\end{document}